\documentclass[10pt,twocolumn,letterpaper]{article}

\usepackage{iccv}
\usepackage{times}
\usepackage{epsfig}
\usepackage{graphicx}
\usepackage{amsmath}
\usepackage{amssymb}

\usepackage{helvet}  
\usepackage{courier}  
\usepackage[hyphens]{url}  
\usepackage{caption} 

\usepackage{algorithm}
\usepackage{algorithmic}

\usepackage{multirow}
\usepackage{multicol}
\usepackage{array}
\usepackage{colortbl}
\usepackage{booktabs}
\usepackage{color} 
\usepackage{color}

\usepackage[breaklinks=true,bookmarks=false,colorlinks]{hyperref}

\iccvfinalcopy 

\def\iccvPaperID{****} 

\ificcvfinal\fi

\begin{document}

\title{Towards High-Quality Temporal Action Detection with Sparse Proposals}

\author
{
Jiannan Wu$^{1}$, 
~~~
Peize Sun$^{1}$, 
~~~
Shoufa Chen$^{1}$,
\\
Jiewen Yang$^{2}$, 
~~~
Zihao Qi$^{2}$, 
~~~
Lan Ma$^{2,3}$,
~~~
Ping Luo$^{1}$
\\[0.2cm]
${^1}$The University of Hong Kong ~~~
${^2}$TCL Corporate Research~~~
\\
${^3}$HKU-TCL Joint Research Centre for Artificial Intelligence
}


\maketitle
\ificcvfinal\thispagestyle{empty}\fi

\begin{abstract}

Temporal Action Detection (TAD) is an essential and challenging topic in video understanding, aiming to localize the temporal segments containing human action instances and predict the action categories. The previous works greatly rely upon dense candidates either by designing varying anchors or enumerating all the combinations of boundaries on video sequences;
therefore, they are related to complicated pipelines and sensitive hand-crafted designs. Recently, with the resurgence of Transformer, \textit{query-based} methods have tended to become the rising solutions for their simplicity and flexibility. However, there still exists a performance gap between \textit{query-based} methods and well-established methods. In this paper, we identify the main challenge lies in the large variants of action duration and the ambiguous boundaries for short action instances; nevertheless, quadratic-computational global attention prevents query-based methods to build multi-scale feature maps. Towards high-quality temporal action detection, we introduce Sparse Proposals to interact with the hierarchical features. In our method, named SP-TAD, each proposal attends to a local segment feature in the temporal feature pyramid. The local interaction enables utilization of high-resolution features to preserve action instances details. Extensive experiments demonstrate the effectiveness of our method, especially under high tIoU thresholds. \Eg, we achieve the state-of-the-art performance on THUMOS14 (45.7\% on mAP@0.6, 33.4\% on mAP@0.7 and 53.5\% on mAP@Avg) and competitive results on ActivityNet-1.3 (32.99\% on mAP@Avg). Code will be made available at \href{https://github.com/wjn922/SP-TAD}{https://github.com/wjn922/SP-TAD}.

\end{abstract}


\section{Introduction}

\begin{figure}[h]
\begin{center}
   \includegraphics[width=1.0\linewidth]{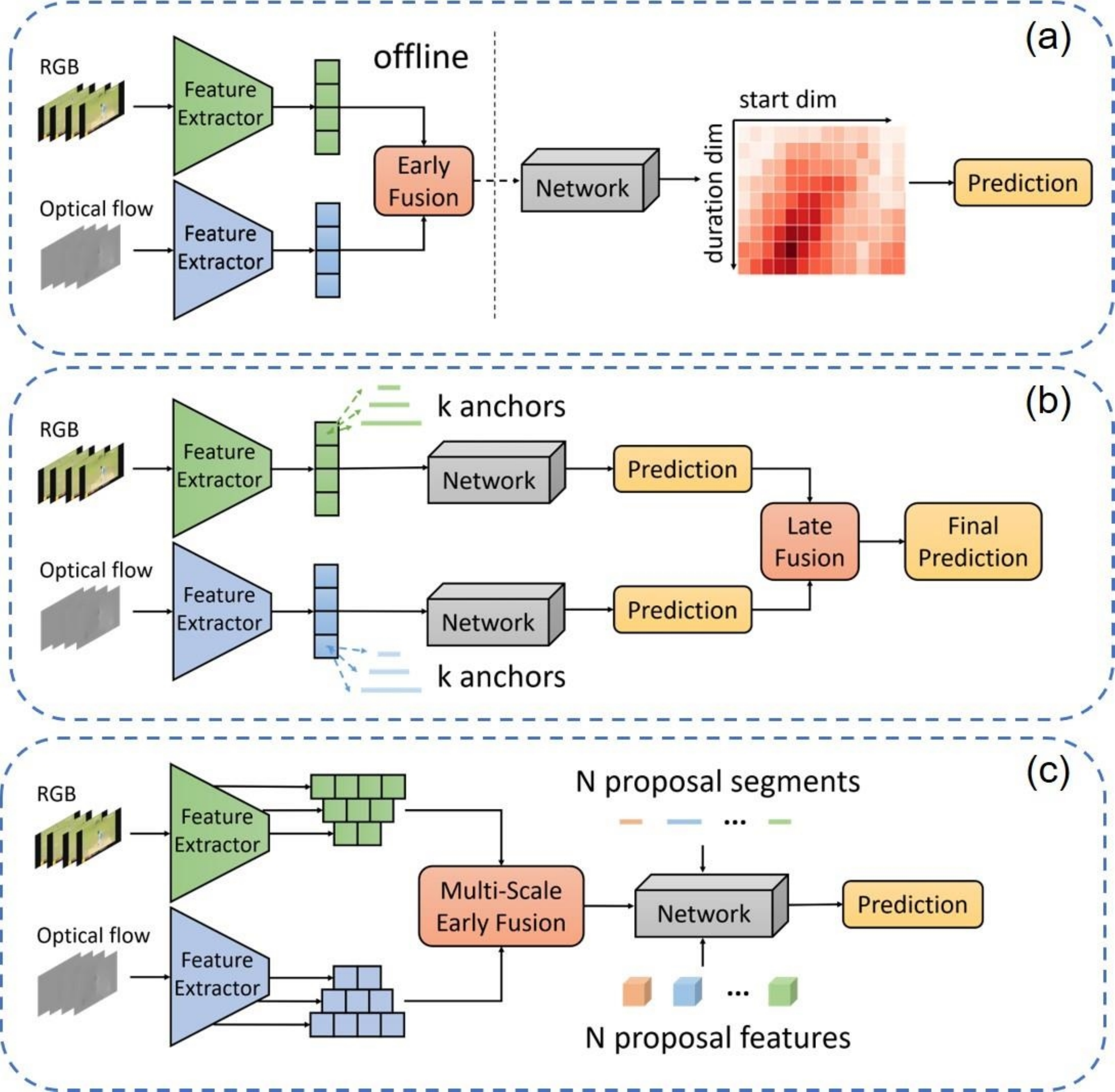}
\end{center}
   \caption{Comparison of current temporal action detection (TAD) pipelines for the two-stream inputs. \textbf{(a)} Offline feature extractor. \textbf{(b)} Two separate networks. \textbf{(c)} Ours. SP-TAD is a unified framework where the two-stream features are fused at the early stage and then adopts only one subsequent network for prediction. The whole network is optimized end-to-end from raw two-stream inputs.}
\label{fig:compare}
\end{figure}


As video resources are growing rapidly in the real world, the task of temporal action detection (TAD) has received considerable critical attention. Given a long untrimmed video, the goal is to locate the temporal segments of person occurrence and predict the human action categories, which serves as a crucial technique in video understanding. 

Previous works have established two well-developed method types and shown impressive performance in this task. (1) \textit{Anchor-based} methods ~\cite{lin2017ssad, chao2018talnet}. They first design multi-scale anchors on every grid of feature sequence. Then, the network performs action classification and boundary regression on these candidates. Since the duration of ground-truth action instances varies dramatically in different videos, those methods suffer from huge computational consumption for placing dense proposal candidates and might have imprecise temporal boundaries. (2) \textit{Boundary-based} methods ~\cite{lin2019bmn, xu2020gtad}. These methods tackle the inaccurate boundary problem in a bottom-up fashion, wherein each matching pair of the video sequence is evaluated. They discard the process of regression and directly generate the confidence scores for the densely distributed proposals. Nevertheless, they can only be used for temporal action proposal generation, thus requiring an external classifier for action classification.

The two well-established methods have been consistently improved and demonstrate their effectiveness with superior performance. However, they still have some limitations. First, these two kinds of methods largely depend on the dense proposal candidates, which would incur a heavy computational burden as the video gets longer. Second, they are vulnerable to artificial parameters, such as the anchor design and confidence threshold. Recently, the \textit{query-based} methods ~\cite{tan2021rtdnet, liu2021tadtr} have raised great interest in the research community. They only employ a small set of queries and thus the network has a simple pipeline as well as gets rid of hand-crafted design. In this paper, we propose a simple and effective framework towards high-quality temporal action detection with sparse proposals (SP-TAD), also belonging to the \textit{query-based} family. The pipeline of our method is shown in Figure \ref{fig:compare} (c). 



In temporal action detection (TAD) task, we identify that an important issue lies in that the duration of action instances in a video varies drastically from several seconds to minutes, and it is difficult for the network to detect short instances. Feature pyramid network (FPN) ~\cite{lin2017fpn} has been widely used in image object detection to solve the problem of large object scale variation. The recent \textit{query-based} method RTD-Net ~\cite{tan2021rtdnet} adopts global attention between the query feature and the global encoded feature, the quadratic-computational complexity prevents it from building multi-scale features. Some other works \cite{gao2020rapnet, wang2021taotad, lin2021afsd} construct the temporal feature pyramid network (TFPN) to mitigate the difficulty of temporal boundary localization. Nonetheless, they all build the TFPN upon the feature extracted from the last layer of the backbone, which contains the high-level representation of the video clip. The down-sample operation in TFPN architecture would further lose the information of short action instances, making it hard for accurate temporal boundary regression. In this work, we present a novel \textit{sparse interaction} process where the proposal feature only interacts with the corresponding local segment feature. Therefore, we can directly use the intermediate layer outputs from backbone to construct the hierarchical feature maps. As the intermediate features have a higher temporal resolution so as to preserve the details of action instances with large variant duration, we show that it is a key element leading to high-quality temporal action detection.

The main contributions of this work are as follows.

\begin{itemize}
\item We propose a simple and unified framework for temporal action detection. Given the RGB frames and optical flows, the whole network is optimized end-to-end from the raw two-stream inputs and outputs the predictions without late fusion. Our method adds a new member to the \textit{query-based} family, which enjoys far fewer candidates and gets rid of the complex hand-crafted designs.
\item We present the \textit{sparse interaction} between the proposal feature and corresponding segment feature. This local interaction enables utilization of high-resolution features which are output by the intermediate layers of backbone. It plays a critical role in high-quality temporal action detection and fast inference speed.
\item Extensive experiments show that SP-TAD outperforms the existing state-of-the-art methods on THUMOS14 and achieves competitive performance on ActivityNet-1.3.
\end{itemize} 

\section{Related Work}

\subsection{Action Recognition}

With  the  rise  of  deep  learning,  action  recognition  has been rapidly developed from the combination of 2D convolutional networks and recurrent neural networks ~\cite{donahue2015rnn} to deep 3D convolutional networks ~\cite{tran2015c3d}. Current deep learning methods can be categorized into two groups: two-stream networks and 3D convolutional networks. The former ones ~\cite{simonyan2014twostream, wang2016tsn, feichtenhofer2019slowfast} adopt two branches to efficiently integrate spatial and temporal information. The latter ones ~\cite{tran2018r3d, xie2018s3d, carreira2017i3d} have a strong capacity to extract the compact spatial-temporal representations simultaneously. All these methods aim to generalize a model with strong robustness and establish correlation and hierarchy of spatial-temporal features in video stream. They are initially developed for action recognition and also served as the backbone for video feature extraction in various video tasks, \eg, temporal action detection and video captioning.

\subsection{Temporal Action Detection}

The pioneering works mainly follow two paradigms: (1) \textbf{Two-stage scheme.} These methods ~\cite{lin2018bsn, lin2019bmn, xu2020gtad, gao2020rapnet} first generate a set of ranked temporal proposals and then classify each proposal for the action category. Most works of this stream are focused on improving the quality of the generated proposals. BSN ~\cite{lin2018bsn} pinpoints local temporal boundaries with high probabilities and evaluates their global confidences. BMN ~\cite{lin2019bmn} develops an end-to-end training pipeline and a boundary-matching strategy for confidence evaluation. DBG ~\cite{lin2020dbg} takes a step forward from BSN by implementing boundary classification and action completeness regression to check through the densely distributed proposals. Some other works ~\cite{zhao2017ssn, zeng2019pgch, qing2021tcanet} take the generated proposals in the first stage as input for further boundary refinement and accurate action classification. However, these two-stage methods go through multiple training stages which have limited correlation, thus may lead to a sub-optimal solution. (2) \textbf{One-stage scheme.} Analogous to object detection, R-C3D ~\cite{xu2017rc3d} and TAL-Net ~\cite{chao2018talnet} adopt the Faster-RCNN ~\cite{ren2015fastercnn} like architecture and accomplish proposal generation and action classification in one network. SSAD ~\cite{lin2017ssad} skips the process of proposal generation and introduces 1D temporal convolution to generate multiple anchors for temporal action detection. A2Net~\cite{yang2020a2net} and AFSD ~\cite{lin2021afsd} visit the anchor-free mechanism, where the network predicts the distance to the temporal boundaries for each temporal location in the feature sequence. AFSD also proposes a novel boundary refinement strategy for precise temporal localization. The one-stage methods enjoy a simpler framework and less hype-parameters tuning.

\subsection{Transformer in Computer Vision}

Transformer~\cite{vaswani2017transformer} was first introduced for natural language processing (NLP) tasks and has raised great attention in the computer vision (CV) community recently ~\cite{dosovitskiy2020vit, wang2021pvt}. The self-attention mechanism in Transformer can dynamically generate global attention weights, which overcomes the disadvantages of convolutional neural networks (CNNs) for local modeling. Transformer has revealed its strong capacity for temporal action proposal generation task. ATAG ~\cite{chang2021atag} proposes an augmented Transformer to capture the long-range temporal contextual information of videos and results in better snippet-level actionness prediction. TAPG Transformer ~\cite{wang2021tapgtr} adopt two complementary Transformer blocks to accomplish the task. One is used to predict the precise boundary scores, and the other learns the rich inter-proposal relationship for reliable proposal confidence evaluation. 

Benefiting from Transformer, DETR ~\cite{carion2020detr} employs a series of object queries instead of anchors as candidates and opens a new view for object detection. Inspired by this work, \textit{query-based} method has become the rising solution for temporal action detection because of its simplicity and flexibility. RTD-Net ~\cite{tan2021rtdnet} leverages the Transformer decoder in which the queries interact with the global encoded feature for directly performing temporal action proposal generation. TadTR ~\cite{liu2021tadtr} restricts the queries to attend a small set of key snippets instead of all the snippets to improve the efficiency.

\section{Approach}

\begin{figure*}[ht]
\begin{center}
\includegraphics[width=0.9\textwidth]{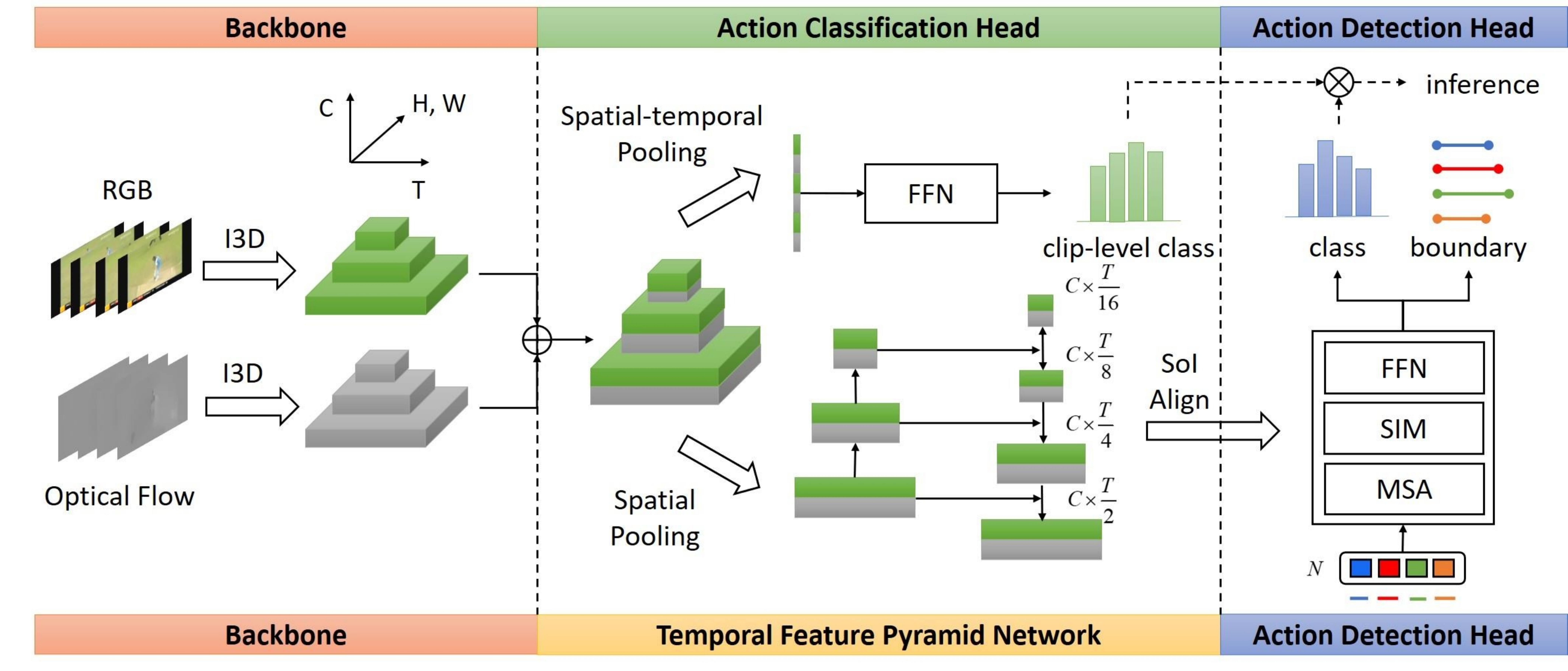}
\end{center}
\caption{The overall architecture of SP-TAD. It mainly consists of backbone networks, a temporal feature pyramid network, an action detection head and an action classification head. During training, the bipartite matching is performed between predicted proposals and ground-truths. During inference, the action detection head predicts a set of $N$ action instances, and the class probabilities are adjusted by the video/clip-level classification scores produced by the action classification head.}
\label{fig:overall}
\end{figure*}


\paragraph{Overview.} Suppose a long untrimmed video $V$ contains $N_{g}$ human action instances and the action instance set is represented by $\mathbf{\Psi}_{g}=\left \{ \hat{\psi} _{n}=\left ( \hat{t}_{n}^{s}, \hat{t}_{n}^{e}, \hat{c}_{n} \right ) \right \}_{n=1}^{N_{g}}$, where $\hat{t}_{n}^{s}$, $\hat{t}_{n}^{e}$ and $\hat{c}_{n}$ represent the start timestamp, end timestamp and action category of the $n$-th action instance, respectively. The goal of temporal action detection (TAD) is to predict a set $\mathbf{\Psi}_{p}$ containing the action proposals with the action classes. $\mathbf{\Psi}_{p}$ should locate the ground-truth action instances precisely and predict the action categories correctly. To achieve this goal, we propose a simple and unified architecture named SP-TAD. As illustrated in Figure \ref{fig:overall}, SP-TAD is an end-to-end training framework for the two-stream inputs. It mainly consists of four parts: 3D backbone networks, a temporal feature pyramid network, an action detection head and an action classification head. 

Specifically, for a given video, we first generate the video clips with the shape $(3, T, H, W)$ and extract the corresponding optical flows. SP-TAD receives the two-stream inputs of a video and extracts the 3D features from them separately. The intermediate layer outputs of backbone networks are used to build the 3D hierarchical feature maps. Afterward, we perform average spatial pooling on the concatenated 3D features to get the 1D sequences and construct a temporal feature pyramid network upon them. The 1D multi-scale features are fed into the action detection head to obtain the temporal segments and action categories results. For the action classification head, it receives the spatial-temporal pooled features from 3D hierarchical feature maps as input and predicts the clip-level action classes probabilities. During training, we apply bipartite matching to create a unique alignment ~\cite{carion2020detr} between the predicted results and ground-truth action instance. The whole network is optimized end-to-end using the combination of set prediction loss ~\cite{carion2020detr} and action classification loss.

\subsection{SP-TAD}

\subsubsection{Backbone}

We adopt the widely-used I3D network ~\cite{carreira2017i3d} as our feature extractor for both RGB frames and optical flows, as it has the ability to model temporal information and proves its superior performance in action recognition. The last three stage outputs from I3D network are extracted to build the three-level hierarchical 3D feature maps. For a video clip with shape $(C, T, H, W)$, the temporal strides and spatial strides of the three-level features are [2, 4, 8] and [8, 16, 32], respectively. Finally, the same level features of RGB and flow streams are concatenated along the channel dimension.

\subsubsection{Temporal Feature Pyramid Network} 

After obtaining the three-level 3D feature maps from the backbone network, we first perform average spatial pooling to get the three-level 1D feature sequences. The features are further transformed by 1D convolutional layers with the output channel as 256. Then, the semantic information of high-level features is passed to low-level features in a top-down fashion. And a 1D max-pooling layer with stride as 2 is added upon the highest level feature. In this way, we construct a four-level feature pyramid with the minimal temporal resolution as $T/16$. Since the lower level features have a higher temporal resolution, they are responsible for predicting short action instances.

\subsubsection{Action Detection Head}

The action detection head receives the multi-level features generated by the temporal feature pyramid network (TFPN) and then predicts the temporal segments and action categories of action instances. The key elements of action detection head design are $N$ learnable \textit{proposal segments} and corresponding \textit{proposal features}. We set the number of proposal candidates as a small value (\eg, $N=50$), which should be larger than the maximum ground-truth action instances number for all videos clips in a dataset. The \textit{proposal segments} are 2-d parameters, indicating the normalized center location and duration of a temporal segment. These \textit{proposal segments} could be set as any size and placed randomly over the feature sequence during initialization, avoiding sophisticated candidate proposal design. The \textit{proposal feature} encodes the rich instance information for each proposal candidate, playing a similar role as the object query in DETR ~\cite{carion2020detr}. The main distinction lies in that every proposal feature only interacts with the corresponding Segment-of-Interest (SoI) feature instead of the global encoded feature, which could be seen as local attention. 

In the action detection head, the proposal features first pass through the multi-head self-attention (MSA) module ~\cite{vaswani2017transformer} to model the relations between each other and then are fed into the sparse interaction module (SIM). 

\begin{figure}[t]
\begin{center}
   \includegraphics[width=1.0\linewidth]{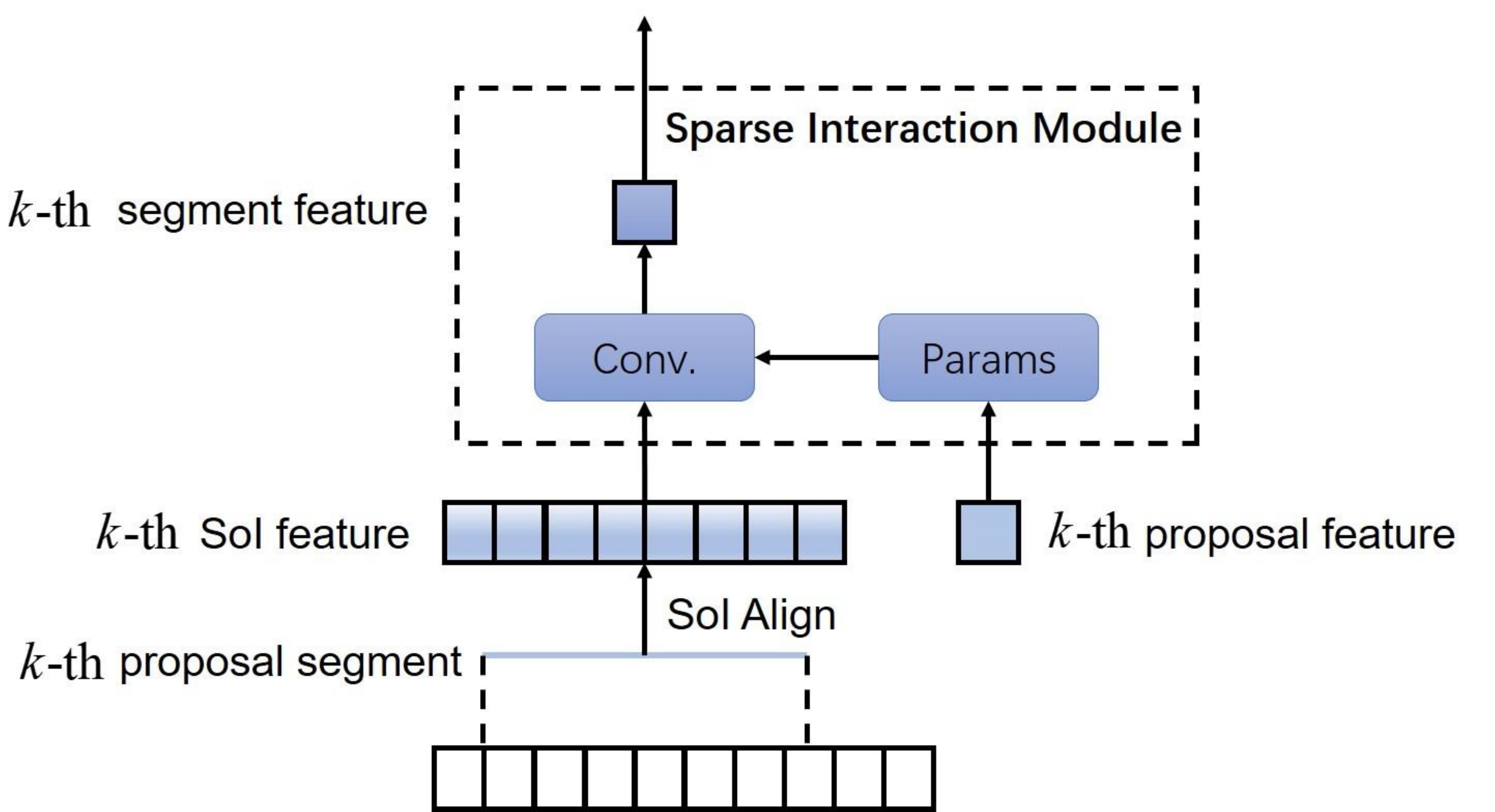}
\end{center}
   \caption{Sparse Interaction Module. For the $k$-th proposal, it performs convolution computation between the proposal feature and the corresponding SoI feature.}
\label{fig:sim}
\end{figure}

Figure \ref{fig:sim} illustrates the \textit{sparse interaction} between the $k$-th proposal feature and corresponding SoI feature in the sparse interaction module. Firstly, due to the ambiguous temporal boundary of an action instance, for the original proposal segment with the region $\left [ t_{s}^{*}, t_{e}^{*} \right ]$, it will be expanded to $\left [ t_{s}^{*}- t_{l}^{*}/\eta , t_{e}^{*}+ t_{l}^{*} / \eta  \right ]$, where $t_{l}^{*}=t_{e}^{*}-t_{s}^{*}$ and $\eta$ is empirically as 5. Then, we use the expanded proposal segment to extract the SoI feature $f_{k} \in \mathbb{R}^{T^{'} \times d}$ from temporal feature pyramid via 1D SoI Align operation ~\cite{xu2020gtad}, where $T^{'}$ is the aligned temporal resolution. For a proposal segment which represents a temporal length of $t$, the corresponding SoI feature is extracted from the $l$-th level feature determined by:

\begin{equation}
    \label{fpn_l}
    l = \left \lfloor l_{0} + log_{2}\left ( t / T \right ) \right \rfloor
\end{equation}

\noindent where $l_{0}$ is the highest level index, and $T$ is the temporal length of the input video clip.

Meanwhile, the proposal feature $p_{k} \in \mathbb{R}^{d}$ produces two parameters with size $d \times d_{h}$ and $d_{h} \times d$ via linear transformation. Afterward, the extracted SoI feature performs matrix multiplication with these two parameters successively. Therefore, the interaction process can be seen as the SoI feature propagating through two 1D convolutional layers. This design helps our model can fully exploit the high-resolution features from the backbone intermediate layers due to its efficiency.

The $N$ outputs features from SIM, called segment features, are further fed into the two-layer feed-forward network (FFN) to obtain the final segment representation. Lastly, two parallel branches are built upon the action detection head to get the final classification scores and boundary regression predictions of $N$ action instances. The classification branch is a linear layer with Sigmoid activation for predicting the probability of each action class. And the regression branch consists of a 3-layer feed-forward network with ReLU activation for temporal boundary regression. 

An important strategy used in the action detection head is the \textbf{iterative refinement}, which is a commonly-used technique ~\cite{cai2018cascadercnn, carion2020detr, zhu2020deformabledetr, sun2020sparsercnn} to improve the performance. We stack the action detection heads and get the predictions of each head. The predicted proposal segments and segment features at each stage serve as the initial proposal segments and proposal features for the next stage so as to be consistently refined. The auxiliary loss is added at intermediate stages to stabilize the training process. We show here the iterative refinement is also the key element towards high-quality temporal action detection. 

\subsubsection{Action Classification Head}

Action recognition ~\cite{wang2016tsn, carreira2017i3d} has been comprehensively studied for a long period and I3D network achieves fairly high accuracy in predicting the action labels of a video clip. Thus, we add the action classification head to predict the video/clip-level action classes probabilities so as to further adjust the classification scores of predicted action instances. Specifically, we directly perform the average spatial-temporal pooling on the three-level 3D features output by the backbone network and feed the pooled feature to a linear layer with Sigmoid activation. The training targets are all action categories happening in the video clip.

\subsection{Training and Inference}

\subsubsection{Training Loss Function}

The predicted action instances set $\mathbf{\Psi}_{p}$ generated by the action detection head contains $N$ samples, which is larger than the maximum number of ground-truth action instances in a dataset. As with DETR ~\cite{carion2020detr}, we expand the target set $\mathbf{\Psi}_{g}$ to the size $N$ by padding $\varnothing$ which indicates no action instance. Then, we adopt the set prediction loss on these two fixed-size sets. And we use the binary cross entropy loss for the action classification head. The overall training loss to optimize the whole network is constituted by these two parts:

\begin{equation}
    \label{total_loss}
    \mathcal{L} = \underbrace{\lambda_{cls}  \cdot  \mathcal{L}_{\mathit{cls}} + \lambda_{L1} \cdot \mathcal{L}_{\mathit{L1}} +
    \lambda_{giou} \cdot \mathcal{L}_{\mathit{giou}} }_{set \ prediction \ loss} + \underbrace{\lambda_{act} \cdot \mathcal{L}_{\mathit{act}}}_{action \ loss}
\end{equation}

\noindent where $\lambda_{cls}$, $\lambda_{L1}$, $\lambda_{giou}$ and $\lambda_{act}$ are weight coefficients. 

The first part of Equation (\ref{total_loss}) is the set prediction loss. It first finds the optimal bipartite matching between predictions and ground-truth action instances by optimizing the matching cost. This provides a one-to-one label assignment pattern, \ie, only those predictions matched with the ground-truth action instances are denoted as \textit{positive samples}. Therefore, for the $n$-th prediction $\psi_{n}$, there exists a unique instance $\hat{\psi}_{\sigma(n)}$ in target set to align with it. Finally, the set prediction loss is computed by the weighted sum of classification loss and regression loss. $\mathcal{L}_{\mathit{cls}}$ is the focal loss ~\cite{lin2017focal} for action classification of action instances. $\mathcal{L}_{\mathit{L1}}$ and $\mathcal{L}_{\mathit{giou}}$ are L1 loss and generalized tIoU loss ~\cite{rezatofighi2019giouloss} between the normalized segment locations of predicted samples and ground-truth instances. They together contribute to the boundary regression.

\begin{equation}
\begin{aligned}
    \label{reg_loss}
    &\mathcal{L}_{\mathit{L1}} = \frac{1}{N_{pos}} \sum_{n: \sigma (n) \neq \varnothing } \left ( \left \| t_{n}^{s}-\hat{t}_{\sigma (n)}^{s} \right \|_{1} + \left \| t_{n}^{e}-\hat{t}_{\sigma (n)}^{e} \right \|_{1} \right )  , \\ 
    &\mathcal{L}_{\mathit{giou}} = \frac{1}{N_{pos}} \sum_{n: \sigma (n) \neq \varnothing } \left ( 1 - GIoU\left ( \psi_{n}, \hat{\psi}_{\sigma (n)} \right ) \right ) 
\end{aligned}
\end{equation}

\noindent where $N_{pos}$ is the total number of positive samples.

\subsubsection{Inference} The inference process of the SP-TAD is simple. Given a video, $N$ predicted action instances associated with the classification probabilities and temporal segments are retrieved from the last action detection head. The classification probabilities are further adjusted by multiplying the video/clip-level action classes scores produced by the action classification head.

\section{Experiments}

\begin{table*}[t]
\begin{center}
{\begin{tabular}{l | l | c |c | c c c c c c}

\toprule

Type & Method & Backbone & One-stage & 0.3 & 0.4 & 0.5 & 0.6 & 0.7 & Avg \\

\arrayrulecolor{white}\hline
\arrayrulecolor{black}\hline
\arrayrulecolor{white}\hline

\multirow{8}{*}{Anchor-based} & SSAD\cite{lin2017ssad} & TS & \checkmark & 43.0 & 35.0 & 24.6 & - & - & - \\
 & TURN\cite{gao2017turn} & C3D & \checkmark & 44.1 & 34.9 & 25.6 & - & - & - \\
 & R-C3D\cite{xu2017rc3d} & C3D & \checkmark & 44.8 & 35.6 & 28.9 & - & - & - \\
 & CBR\cite{gao2017cbr} & TS & \checkmark & 50.1 & 41.3 & 31.0 & 19.1 & 9.9 & 30.3 \\
 & TAL-Net\cite{chao2018talnet} & I3D & \checkmark & 53.2 & 48.5 & 42.8 & 33.8 & 20.8 & 39.8 \\
 & GTAN\cite{long2019gtan} & P3D & \checkmark & 57.8 & 47.2 & 38.8 & - & - & - \\
 & PBRNet\cite{liu2020pbrnet} & I3D & \checkmark & 58.5 & 54.6 & 51.3 & 41.8 & 29.5 & 47.1 \\
 & AFSD\cite{lin2021afsd} & I3D & \checkmark & 67.3 & 62.4 & 55.5 & 43.7 & 31.1 & 52.0 \\
 
\arrayrulecolor{white}\hline
\arrayrulecolor{black}\hline
\arrayrulecolor{white}\hline

\multirow{5}{*}{Boundary-based} & SSN\cite{zhao2017ssn} & TS & & 51.0 & 41.0 & 29.8 & - & - & - \\
 & BSN\cite{lin2018bsn} & TS & & 53.5 & 45.0 & 36.9 & 28.4 & 20.0 & 36.8 \\
 & BMN\cite{lin2019bmn} & TS & & 56.0 & 47.4 & 38.8 & 29.7 & 20.5 & 38.5 \\
 & DBG\cite{lin2020dbg} & TS & & 57.8 & 49.4 & 42.8 & 33.8 & 21.7 & 41.1 \\
 & G-TAD\cite{xu2020gtad} & TS & & 54.5 & 47.6 & 40.2 & 30.8 & 23.4 & 39.3 \\
 & BU-TAL\cite{zhao2020butal} & I3D & & 53.9 & 50.7 & 45.4 & 38.0 & 28.5 & 43.3 \\
 
\arrayrulecolor{white}\hline
\arrayrulecolor{black}\hline
\arrayrulecolor{white}\hline

\multirow{3}{*}{Query-based} & RTD-Net\cite{tan2021rtdnet} & I3D & & 58.5 & 53.1 & 45.1 & 36.4 & 25.0 & 43.6 \\
 & TadTR\cite{liu2021tadtr} & I3D & \checkmark & 62.4 & 57.4 & 49.2 & 37.8 & 26.3 & 46.6 \\
 & \textbf{SP-TAD (Ours)} & I3D & \checkmark & \textbf{69.2} & \textbf{63.3} & \textbf{55.9} & \textbf{45.7} & \textbf{33.4} & \textbf{53.5} \\

\arrayrulecolor{white}\hline
\arrayrulecolor{black}\hline
\arrayrulecolor{white}\hline
\end{tabular}}
\end{center}
\vspace{-2mm}
\caption{Comparison with the state-of-the-art methods on THUMOS14 testing set. One-stage means the method does not need an external classifier for action category prediction.}
\label{table:thumos}
\vspace{-4mm}
\end{table*}

\subsection{Datasets and Setup}


\paragraph{Datasets.} We conduct our experiments on two popular datasets: \textbf{THUMOS14} ~\cite{jiang2014thumos} dataset contains 1,010 and 1,574 untrimmed videos with 200 class categories in validation and testing set, respectively. For the temporal action localization task, there are 200 validation videos and 213 testing videos with temporal annotation of 20 action classes. \textbf{ActivityNet-1.3} ~\cite{caba2015activitynet} dataset contains 19,994 untrimmed videos of 200 action categories; all videos are temporally annotated. Additionally, they are divided into training, validation and testing set by the ratio of 2:1:1.


\paragraph{Implementation Details.} On THUMOS14 dataset, we encode the videos at 10 frames per second (fps) with the spatial resolution as $112 \times 112$. Following the previous practice ~\cite{lin2019bmn, lin2021afsd}, we use the sliding windows to generate consecutive video clips. As the 98\% action instances of the dataset are less than 25.6s, we set the temporal length of each clip as 256 frames. The temporal strides between the adjacent video clips are set as 30 and 128 during training and inference phase, respectively. Because the adjacent clips have overlaps, which would produce redundant predictions, we adopt the soft-NMS ~\cite{bodla2017softnms} as post-process. For ActivityNet-1.3, as each video only contains 1.5 action instances on average, we encode each video to the fixed temporal length of 768 frames. On both datasets, random crop and horizontal flipping are applied as the image level data augmentation during training. The spatial size of the cropped image is set as $96 \times 96$. The number of learnable proposals on each sample is selected as 50 for THUMOS14 and 100 for ActivityNet-1.3.

The I3D backbone networks are initialized with the parameters pre-trained on Kinetics dataset ~\cite{kay2017kinetics400}. We use the AdamW ~\cite{loshchilov2018adamw} optimizer with the weight decay of 0.0001 to train the network and we set the batch size as 16. The learning rate is 0.0001 for the first 12 epochs and 0.00001 for the following 4 epochs.


\paragraph{Evaluation Metrics.} For the temporal action detection task, mean Average Precision (mAP) is calculated as the evaluation metric. On ActivityNet-1.3, mAP with tIoU thresholds $\left \{ 0.5,0.75,0.95 \right \}$ and average mAP under tIoU thresholds $\left [ 0.5 : 0.05: 0.95 \right ]$ are reported. On THUMOS14, we compute the mAP with tIoU thresholds $\left [ 0.3:0.1:0.7 \right ]$ and average mAP.

\subsection{Main Results}

\paragraph{THUMOS14} The comparison results of the state-of-the-art methods on THUMOS14 are summarized in Table \ref{table:thumos}. It can be seen that SP-TAD outperforms all the previous methods by a large margin, especially under high tIoU thresholds. For instance, SP-TAD achieves significant improvement from 43.7\% to 45.7\% on mAP@0.6 and from 31.1\% to 33.4\% on mAP@0.7 compared with the previous best method AFSD ~\cite{lin2021afsd}. It should be noted that AFSD has already been far ahead of the performance over other methods. Our model only employs a set of sparse queries to achieve the remarkable performance while AFSD uses dense anchors as candidates. When compared with the two-stage methods which generate proposals first and then classify the action categories, SP-TAD not only has a simple and unified framework, but also shows superior performance over them.

\begin{table}[t]
\begin{center}
{\begin{tabular}{l | c c c c}

\toprule

Method & 0.5 & 0.75 & 0.95 & Avg \\

\arrayrulecolor{white}\hline
\arrayrulecolor{black}\hline
\arrayrulecolor{white}\hline

\multicolumn{5}{l}{\textbf{Anchor-based}} \\

\arrayrulecolor{white}\hline
\arrayrulecolor{black}\hline
\arrayrulecolor{white}\hline

TAL-Net\cite{chao2018talnet} & 38.23 & 18.30 & 1.30 & 20.22 \\
SSAD\cite{lin2017ssad} & 44.39 & 29.65 & \textbf{7.09} & 29.17 \\
AFSD\cite{lin2021afsd} & \textbf{52.38} & \textbf{35.27} & 6.47 & \textbf{34.39} \\
 
\arrayrulecolor{white}\hline
\arrayrulecolor{black}\hline
\arrayrulecolor{white}\hline

\multicolumn{5}{l}{\textbf{Boundary-based}} \\

\arrayrulecolor{white}\hline
\arrayrulecolor{black}\hline
\arrayrulecolor{white}\hline

SSN\cite{zhao2017ssn} & 39.12 & 23.48 & 5.49 & 23.98 \\
BSN\cite{lin2018bsn} & 46.45 & 29.96 & 8.02 & 30.03 \\
BMN\cite{lin2019bmn} & 50.07 & \textbf{34.78} & 8.29 & 33.85 \\
G-TAD\cite{xu2020gtad} & \textbf{50.36} & 34.60 & 9.02 & \textbf{34.09} \\
BU-TAL\cite{zhao2020butal} & 43.37 & 33.91 & \textbf{9.21} & 30.12 \\
 
\arrayrulecolor{white}\hline
\arrayrulecolor{black}\hline
\arrayrulecolor{white}\hline

\multicolumn{5}{l}{\textbf{Query-based}} \\

\arrayrulecolor{white}\hline
\arrayrulecolor{black}\hline
\arrayrulecolor{white}\hline

RTD-Net\cite{tan2021rtdnet} & 47.21 & 30.68 & \textbf{8.61} & 30.83 \\
TadTR\cite{liu2021tadtr} & 49.08 & 32.58 & 8.49 & 32.27 \\
\textbf{SP-TAD (Ours)} & \textbf{50.06} & \textbf{32.92} & 8.44 & \textbf{32.99} \\
 
\arrayrulecolor{white}\hline
\arrayrulecolor{black}\hline
\arrayrulecolor{white}\hline
\end{tabular}}
\end{center}
\caption{Comparison with the state-of-the-art methods on ActivityNet-1.3 validation set.}
\label{table:activitynet}
\end{table}

\begin{figure}[t]
\begin{center}
   \includegraphics[width=1.0\linewidth]{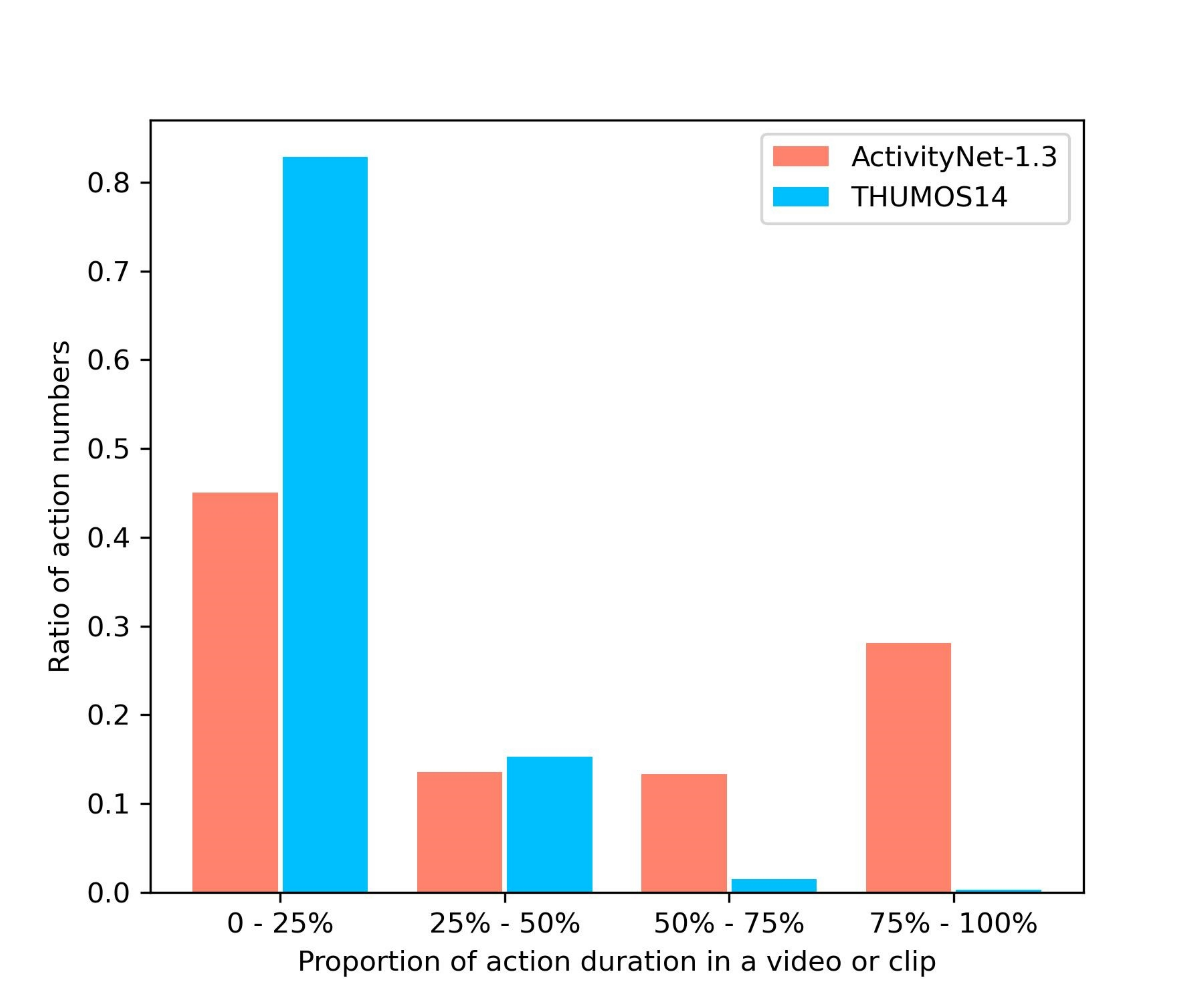}
\end{center}
   \caption{Comparison of action number distribution with respect to the action duration proportion between ActivityNet-1.3 and THUMOS14 datasets.}
\label{fig:action_ratio}
\end{figure}

\paragraph{ActivityNet-1.3} Table \ref{table:activitynet} illustrates the temporal action detection performance of different methods on ActivityNet-1.3 validation set. 
It can be seen that our model outperforms other \textit{query-based} methods on mAP@Avg. On the other hand, we observe that \textit{anchor-based} and \textit{boundary-based} methods have higher performance over the \textit{query-based} methods, this is because the former two methods generate massive candidates and select the top 100 predictions from them, while the last one only predicts 100 proposals or less for each video. It should be noted that \textit{query-based} methods have competitive performance under high tIoU thresholds, indicating that the methods have precise temporal boundaries. 

Moreover, we compare the action number distribution between ActivityNet-1.3 and THUMOS14 in Figure \ref{fig:action_ratio}, where x-axis indicates the percentage of action length in a video or clip and y-axis represents the ratio of action number in a specified interval to all the action number in the dataset. Particularly, on THUMOS14, we first use the sliding windows with an overlap ratio of 0.5 to slide on the video. Then a ground-truth action instance is recorded when its tIoA\footnote{For two temporal segments $p1$ and $p2$, tIoA of $p1$ is defined as $tIoA_{p1}=\frac{p1\bigcap p2}{p1}$.} to the truncated window is larger than 0.5. This is also the strategy for selecting samples during the training phase. It could be seen that the data distribution of these two datasets is quite different. On THUMOS14, over 80\% action instances take up less than 25\% duration time in a video clip and there are hardly any action instances belonging to the long instances; while ActivityNet-1.3 has a more even distribution regarding the duration time of action instances. Thus, it better highlights the advantage of our method.

\subsection{Comparison of Inference Speed}

Thanks to the \textit{sparse interaction} design, our model can not only utilize the high-resolution features to achieve superior performance, but also have high efficiency due to local attention. To verify this statement, we compare the inference speed between our method and other state-of-the-art end-to-end training methods. For a fair comparison, we use the RGB frames as inputs to the network. As shown in Table \ref{table:fps}, our model can process the videos at 5574 FPS on a single V100 GPU, which is much faster than the existing methods. 

\begin{table}[t] 
	\centering
    \begin{tabular}{l | c | c}

\toprule

Method & GPU & FPS \\

\arrayrulecolor{white}\hline
\arrayrulecolor{black}\hline
\arrayrulecolor{white}\hline

SS-TAD \cite{buch2019sstad} & TITAN Xm & 701 \\
R-C3D \cite{xu2017rc3d} & TITAN Xp & 1030 \\
PBRNet \cite{liu2020pbrnet} & 1080Ti & 1488 \\
AFSD \cite{lin2021afsd} & 1080Ti & 3259 \\
AFSD \cite{lin2021afsd} & V100 & 4057 \\

\arrayrulecolor{white}\hline
\arrayrulecolor{black}\hline
\arrayrulecolor{white}\hline

\textbf{SP-TAD (Ours)} & V100 & \textbf{5574} \\

\arrayrulecolor{white}\hline
\arrayrulecolor{black}\hline
\arrayrulecolor{white}\hline
\end{tabular}
	\caption{Comparison of inference speed with state-of-the-art end-to-end training methods.}
    \label{table:fps}
\end{table}

\subsection{Ablation Study}

In this section, we conduct several ablation studies on THUMOS14, aiming to explore the key elements that make for high-quality temporal action detection.

\subsubsection{The Effect of Unified Backbone} 

Most of the previous works ~\cite{lin2019bmn, xu2020gtad, zhao2020butal, tan2021rtdnet} adopt an offline feature extractor to obtain the video features in advance and then design the network to process the fixed features. We hypothesize that fixed video features would limit the capacity of the current TAD methods because of the misalignment between different datasets. The recent works ~\cite{liu2020pbrnet, lin2021afsd} jointly train the backbone but do not explicitly explore the effect of the unified backbone in the network. 

We gradually increase the learning rate for training backbone while keeping the learning rate for other parts unchanged as $1e^{-4}$. Specifically, when the learning rate of backbone equals 0, the backbone could be seen as an offline feature extractor. From Table \ref{table:ab_backbone}, we observe the consistent improvement by increasing the learning rate, which clearly proves that the unified backbone in the network would benefit the TAD task.

\begin{table}[t] 
	\centering
    \setlength{\tabcolsep}{1.5mm}
\begin{tabular}
{c c c c c c c}
\toprule
Backbone Lr & 0.3 & 0.4 & 0.5 & 0.6 & 0.7 & Avg \\
\midrule
0 & 65.1 & 58.3 & 49.5 & 39.1 & 27.6 & 48.0 \\
$1e^{-6}$ & 65.4  & 59.6  & 51.0  & 40.1  & 28.6 & 48.9 \\
$1e^{-5}$ & 67.2 & 62.1 & 51.8 & 42.1 & 29.7 & 50.6 \\
$1e^{-4}$ & \textbf{69.2} & \textbf{63.3} & \textbf{55.9} & \textbf{45.7} & \textbf{33.4} & \textbf{53.5} \\
\bottomrule
\end{tabular}
	\caption{Effect of the learning rate of backbone. `Lr` means learning rate.}
    \label{table:ab_backbone}
\end{table}

\subsubsection{The Effect of Temporal Feature Pyramid} 

The previous works neglect the features output by the intermediate layers of the backbone. However, we contend that these features preserve rich temporal information and play a key role in high-quality temporal action detection. Here we study different kinds of features as the input to the action detection head: 

\begin{itemize}
    \item ``Single Level'' means only using the last layer output of the backbone.
    \item ``High Level TFP'' represents that building the four-level temporal feature pyramid upon the highest level features from the backbone.
    \item ``Inter Level TFP'' is the four-level feature pyramid used in our model, which uses the intermediate layer outputs of the backbone to construct the hierarchical feature maps.
\end{itemize}

From Table \ref{table:ab_fpn}, we find that ``Single Level'' feature performs even better than the ``High Level TFP''. This is because most of the aligned SoI features are extracted from the higher levels of temporal feature pyramid and they have lower temporal resolution than the ``Single Level'' feature. These high-level features would lose the details of short action instances and thus degenerate the performance instead. A straightforward but effective solution is using the intermediate outputs from the backbone to construct a temporal feature pyramid. It achieves significant performance gain, especially for high tIoU thresholds, as illustrated in the last row of Table \ref{table:ab_fpn}.

\begin{table}[t]
	\centering
    \setlength{\tabcolsep}{1.5mm}
\begin{tabular}
{c c c c c c c}
\toprule
Features & 0.3 & 0.4 & 0.5 & 0.6 & 0.7 & Avg \\
\midrule
Single Level & 67.5 & 61.2 & 52.7 & 41.8 & 29.4 & 50.2 \\
High Level TFP & 65.8  & 59.9  & 51.5  & 40.0  & 28.1 & 49.0 \\
Inter Level TFP & \textbf{69.2} & \textbf{63.3} & \textbf{55.9} & \textbf{45.7} & \textbf{33.4} & \textbf{53.5} \\
\bottomrule
\end{tabular}
	\caption{Effect of the temporal feature pyramid. `Inter Level TFPN` used in our model gets the highest performance.}
    \label{table:ab_fpn}
\end{table}

\subsubsection{The Effect of Iterative Refinement}

We study the effect of the number of stacked action detection heads in Table \ref{table:ab_head}. The first row of Table \ref{table:ab_head} clearly shows that the model has an extremely poor performance of 43.6\% on mAP@Avg without the iterative refinement process, while the performance would be greatly improved by increasing head number to 2, bringing a gain of 6.0\% on mAP@Avg. This proves the effectiveness and necessity of iterative refinement. It also indicates that a good initialization for the \textit{proposal segments} would reduce the prediction difficulty for the action detection head. As head number gradually increases, the performance gain becomes smaller and it saturates at the stage of 4. The iterative refinement is another important element for high-quality temporal action detection.

\begin{table}[t]
	\centering
    \setlength{\tabcolsep}{1.5mm}
\begin{tabular}
{c c c c c c c}
\toprule
Head & 0.3 & 0.4 & 0.5 & 0.6 & 0.7 & Avg \\
\midrule
1 & 60.1 & 54.4 & 45.7 & 34.7 & 23.1 & 43.6 \\
2 & 66.0  & 59.4  & 55.5  & 41.1  & 30.0 & 49.6 \\
4 & \textbf{69.2} & \textbf{63.3} & \textbf{55.9} & \textbf{45.7} & \textbf{33.4} & \textbf{53.5} \\
6 & 65.4 & 59.4 & 52.1 & 42.3 & 30.4 & 49.9 \\
\bottomrule
\end{tabular}
	\caption{Effect of the iterative refinement strategy.}
    \label{table:ab_head}
\end{table}

\subsubsection{The Effect of Action Classification Head}

The action recognition results have achieved fairly high thanks to the development of strong 3D backbones. Therefore, we add an action classification head to predict the probability distribution of different actions. It could be used to rectify the classification scores predicted by the action detection head during inference phase. As illustrated in Table \ref{table:ab_cls_head}, it improves the mAP@Avg from 52.1\% to 53.5\% on THUMOS14 testing set, proving this strategy is an effective way to correct the classification error by action detection head.

\begin{table}[t]
	\centering
    \setlength{\tabcolsep}{1.5mm}
\begin{tabular}
{c c c c c c c}
\toprule
Cls Head & 0.3 & 0.4 & 0.5 & 0.6 & 0.7 & Avg \\
\midrule
 & 67.2 & 61.6 & 54.7 & 44.3 & 32.6 & 52.1 \\
\checkmark & \textbf{69.2} & \textbf{63.3} & \textbf{55.9} & \textbf{45.7} & \textbf{33.4} & \textbf{53.5} \\
\bottomrule
\end{tabular}
	\caption{Effect of the action classification head. `Cls Head` means action classification head.}
    \label{table:ab_cls_head}
\end{table}

\subsubsection{The Choice of Video Clip Length}

\begin{table}[t]
	\centering
    \setlength{\tabcolsep}{1.5mm}
\begin{tabular}
{c c c c c c c}
\toprule
Clip length & 0.3 & 0.4 & 0.5 & 0.6 & 0.7 & Avg \\
\midrule
224 & 68.5 & 62.6 & 55.0 & 43.6 & 31.7 & 53.3 \\
256 & \textbf{69.2} & \textbf{63.3} & \textbf{55.9} & \textbf{45.7} & \textbf{33.4} & \textbf{53.5} \\
288 & 68.3 & 62.4 & 54.4 & 43.7 & 31.6 & 51.3 \\
320 & 67.2 & 62.1 & 53.4 & 43.1 & 30.7 & 49.9 \\
\bottomrule
\end{tabular}
	\caption{Effect of the video clip length.}
    \label{table:ab_len}
\end{table}

In real applications, a long untrimmed video would contain several action instances with large duration variance. Therefore, applying the sliding windows on the video is a useful technique for accurate temporal action detection. We vary the length of sliding window from 224 to 320 while keeping the overlap ratio as 0.5 during inference. The results are summarized in Table \ref{table:ab_len}. A window with the length of 256 can cover 98\% action instances in all the videos, thus it would be a suitable choice. When the length decreases to 224, the network can still achieve strong performance but would falsely detect the over-long action instances. On the other hand, increasing the length of sliding window means that the action segments would take less proportion in a clip, which makes it harder to detect the short action instances. It is reasonable to expect a slight performance drop by adopting a larger window.

\subsection{Visualization Results}

\begin{figure}[t]
\begin{center}
   \includegraphics[width=1.0\linewidth]{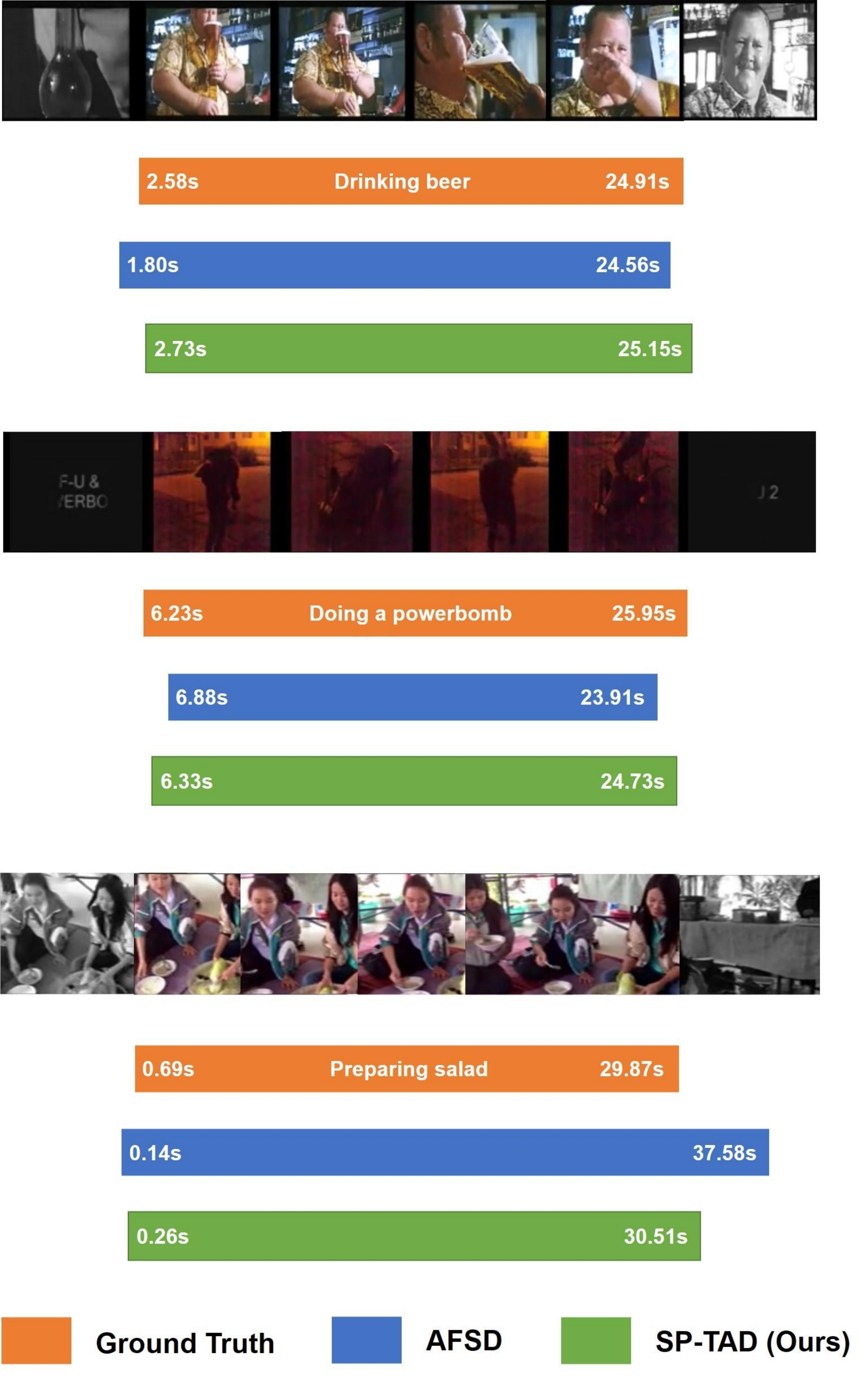}
\end{center}
   \caption{Visualization results of our method and AFSD on ActivityNet-1.3.}
\label{fig:vis_anet}
\end{figure}

In this section, we compare the visualization results of our method and the state-of-the-art method AFSD ~\cite{lin2021afsd} on ActivityNet-1.3. From Figure \ref{fig:vis_anet}, we can observe that the temporal boundaries of our method are more close to the ground-truth action instances. This phenomenon indicates that our method performs better under high tIoU threshold, which is also supported in Table \ref{table:activitynet}.

\section{Conclusion}

In this paper, we present a simple and effective framework for high-quality temporal action detection, named SP-TAD. Our model belongs to the \textit{query-based} family which enjoys the simple pipeline by introducing a small set of learnable proposals and gets rid of the hand-crafted anchor design. It is a unified framework and thus can be optimized end-to-end from raw two-stream inputs. Moreover, we propose the novel \textit{sparse interaction} that enables utilization of high-resolution features, leading to the high performance and fast inference speed. We also identify the key elements for producing high-quality temporal segments: the unified backbone, intermediate outputs form the backbone and the iterative refinement strategy. Experiments demonstrate that our model achieves state-of-the-art performance on THUMOS14 and competitive results on ActivityNet-1.3.

{\small
\bibliographystyle{ieee_fullname}
\bibliography{egbib}
}

\end{document}